\documentclass[conference]{IEEEtran}
\IEEEoverridecommandlockouts
\usepackage{cite}
\usepackage{amsmath,amssymb,amsfonts}
\usepackage{algorithmic}
\usepackage{graphicx}
\usepackage{textcomp}
\usepackage{xcolor}
\usepackage{caption}
\usepackage{subcaption}
\usepackage{float}
\usepackage{balance}
\usepackage{url}
\def\BibTeX{{\rm B\kern-.05em{\sc i\kern-.025em b}\kern-.08em
    T\kern-.1667em\lower.7ex\hbox{E}\kern-.125emX}}
\begin{document}

\title{Understanding Path Planning Explanations}

\author{
\IEEEauthorblockN{Amar Halilovic}
\IEEEauthorblockA{\textit{Ulm University, Ulm, Germany}\\
amar.halilovic@uni-ulm.de
}

\and

\IEEEauthorblockN{Senka Krivic}
\IEEEauthorblockA{\textit{University of Sarajevo, Sarajevo, BiH} \\
senka.krivic@etf.unsa.ba
}
}

\maketitle

\begin{abstract}
Navigation is a must-have skill for any mobile robot.  
A core challenge in navigation is the need to account for an ample number of possible configurations of environment and navigation contexts. 
We claim that a mobile robot should be able to explain its navigational choices making its decisions understandable to humans. 
In this paper, we briefly present our approach to explaining a robot’s navigational decisions through visual and textual explanations.
We propose a user study to test the understandability and simplicity of the robot explanations and outline our further research agenda.
\end{abstract}


\section{Introduction}
There is an increasing deployment of autonomous robots in various domains \cite{breazeal2005socially}.
Albeit many benefits, a lack of transparency and accountability in their decision-making exists \cite{felzmann2019robots}.
The growing tendency to adopt autonomous robots promotes the need for the explainability of their decisions \cite{lim2021social}, which can help increase human understanding and trust in them \cite{leichtmann2023effects}.
Navigation is a pivotal aspect of an autonomous robot decision-making spectrum.
Local decisions in navigation play a key role in achieving accurate and efficient navigation in changing environments. 
They involve a robot's ability to make quick decisions about its movement.
We tackle the challenge of explainability in robotics by explaining a robot's local navigational decisions \cite{karalus2021explanations,halilovic2022explaining,halilovic2023visuo}.

\section{Explainable Navigation}
\begin{figure*}[t]
\centering

\begin{subfigure}[b]{0.24\textwidth}
\centering
\includegraphics[width=\textwidth, height=4.35cm]{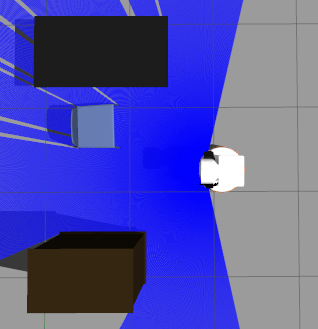}
\caption{Navigation situation}
\label{fig:r1}
\end{subfigure}
\begin{subfigure}[b]{0.24\textwidth}
\centering
\includegraphics[width=\textwidth]{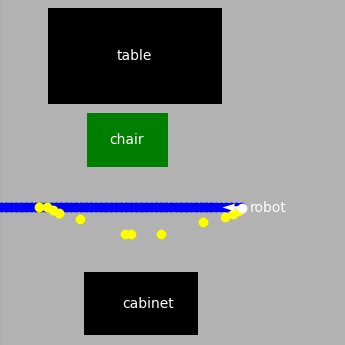}
\caption{Chair-movability expl.}
\label{fig:r2}
\end{subfigure}
\begin{subfigure}[b]{0.24\textwidth}
\centering
\includegraphics[width=\textwidth]{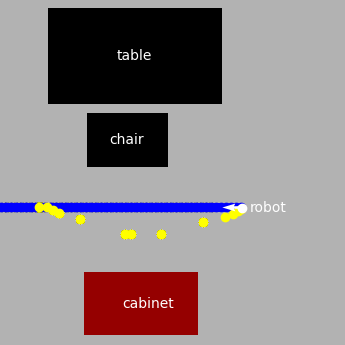}
\caption{Cabinet-movability expl.}
\label{fig:r3}
\end{subfigure}
\begin{subfigure}[b]{0.24\textwidth}
\centering
\includegraphics[width=\textwidth]{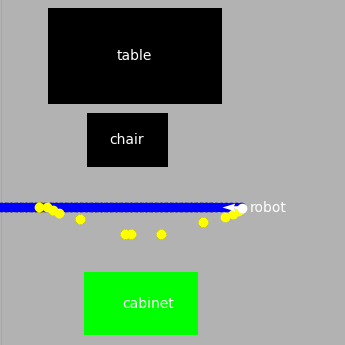}
\caption{Cabinet-openability expl.}
\label{fig:r4}
\end{subfigure}

\caption{
\textbf{Explanation multi-modality coherence \cite{halilovic2023visuo}:}
Fig. \ref{fig:r1} shows a navigation scenario.
The robot TIAGo \cite{pages2016tiago} tries to follow the initial path which leads it to pass between the table and the closed cabinet, but the pathway in front of it is narrowed by a chair.
Instead of going straight, the robot starts deviating to its left following a new, locally created, path to avoid hitting the chair.
Each object is accompanied by its actionable properties (movability, openability, etc.), viz. affordances.
Together with the initial path (blue), local path (yellow), and the robot's position and orientation (white), the objects (colored to the corresponding affordances) constitute explanation maps (see Fig. \ref{fig:r2}, \ref{fig:r3}, and \ref{fig:r4}).
Object-affordance pairs are colored according to their contributions to the robots' deviation from the initial path (green---increase in deviation, red--decrease in deviation).
On one explanation map one object is colored according to one object-affordance pair.
The chair (Fig. \ref{fig:r2}) forces the robot to deviate from the initial path.
The cabinet (Fig. \ref{fig:r3}) limits the size of the deviation, and if it was moved the robot would deviate more.
If the cabinet was open (Fig. \ref{fig:r4}) the robot would not have space to deviate this much, and it would potentially have the strongest influence (most brightly colored) on the change (decrease) in the deviation.
The table is far enough from the robot that it has no effect on its navigation.
Along with the visual explanations come the textual counterfactual explanations and suggestions:
\textbf{\ref{fig:r2})} \textit{"Because of the chair right-front of me, I deviate from the initial plan."} or \textit{"If the chair right-front of me was not there, I would deviate less from the initial plan."};
    \textit{"Dear human, please move the chair, so I can proceed more smoothly."};
\textbf{\ref{fig:r3})} \textit{"If the cabinet left-front of me was not there, I would deviate more from the initial plan."};
\textbf{\ref{fig:r4})} \textit{"If the cabinet left-front of me was open, I would deviate less from the initial plan."}
}

\label{fig:exps}
\end{figure*}

When navigating among humans, robots might be unaware of particular human social norms.
Thus, humans could regard the robots' behaviors as unfitting.
To minimize the potential for misunderstandings, we work on making robot navigation more understandable to humans by utilizing Explainable Artificial Intelligence (XAI) \cite{gunning2019xai}.
With our visual-explanation approach \cite{halilovic2022explaining}, based on LIME \cite{ribeiro2016should}, obstacles in a robot's neighborhood are highlighted, conditional on their importance for the navigational action (the local path) of the robot.
Combining our visual explanations with semantically-rich textual explanations and suggestions  (see Fig. \ref{fig:exps}) we achieve a real-time explanation multi-modality \cite{halilovic2023visuo}. 
Using our multi-modal approach, a robot can explain both visually and textually its reasoning, which is deemed inappropriate by humans, e.g. getting too close to humans when passing.
Additionally, it can suggest a mitigation strategy to get out of an unwanted situation, e.g. \textit{"Dear human, here is a chair on my way. I need to slip through. Could you move it from my way?"}.

\section{Explanation Understanding}
The most researched motion planning explanation approaches in the literature are explanations of planning failures \cite{hauser2014minimum, kwon2018expressing}.
Brandao et al. \cite{brandao2021towards} introduce a taxonomy of motion planning explanation approaches and extend the concept to trajectory-contrastive explanations.
Following their taxonomy, our explanations are trajectory-contrastive, as well as environment- and constraint-based.
Our approach is easily extendable to explaining planning failures, e.g. in situations when the robot is stuck, i.e. can not continue following the initial path, nor can find any feasible alternative path.

For the design of our user study, we will extend and formalize our approach to explain both path planning failures and deviations from the initial trajectory, i.e. to be able to answer trajectory-contrastive questions.
Currently, we use one color schema (red---green) for visual explanations and two types of textual explanations, i.e. descriptions and suggestions.

Our main goal is to test whether end-users understand our explanations.
Following good design practice \cite{hoffman2020primer}, we will create different planning failures and trajectory-contrastive scenarios.
After generating explanations for the created scenarios, we will perturb the explanations in the following way:
\begin{itemize}
    \item perturbations of visual explanations will have different coloring schemes, shapes and textures
    \item perturbations of textual explanations will have different lengths and verbalization \cite{rosenthal2016verbalization,perera2016dynamic}
\end{itemize}
We will design a comparative user study to evaluate the understandability and effectiveness of our multi-modal explanation framework by following recommendations for metrics on Explanation Satisfaction reported by Hoffman et al. \cite{hoffman2018metrics}. 
Explanation Satisfaction is a metric that captures users' understanding of explanations presented to them as a posteriori explanation judgment.
We will let participants choose the most understandable explanation perturbations for both scenarios and rate their explanation satisfaction on chosen perturbations.

\section*{Conclusion and Future Work}
We envisage the navigating TIAGo\footnote{\url{https://pal-robotics.com/robots/tiago/}} with a fully autonomous and social explanation system.
Alongside the planned user study, future work will expand on our current explanation apparatus and include semantically richer explanations, motion planning failures explanations, and focus on explanation conveying strategies, i.e. on explanation timing.

\section*{Acknowledgment}
\textit{I express my gratitude to my late supervisor Jun.-Prof. Dr. Felix Lindner for his invaluable support and mentorship in the first phase of my Ph.D. journey. I also express my appreciation to Asst. Prof. Dr. Senka Krivic for her decision and willingness to supervise me in the continuation of my Ph.D.} - Amar Halilovic.

\balance
\bibliographystyle{IEEEtran}
\bibliography{IEEEabrv,sources}

\end{document}